\documentclass[runningheads]{llncs}
\usepackage[T1]{fontenc}
\usepackage{graphicx}
\usepackage{wrapfig} 

\usepackage{array}      
\usepackage{enumitem}   
\usepackage{booktabs}   

\usepackage{amsmath}    
\usepackage{amssymb}    
\usepackage{graphicx}   
\usepackage{booktabs}   
\usepackage{hyperref}   
\usepackage{xcolor}     
\usepackage{caption}    
\usepackage{subcaption} 
\usepackage{cite}       
\usepackage{multirow}   
\usepackage{tikz}

\usepackage{times}
\usepackage{latexsym}
\usepackage{color}
\usepackage{bbding}
\usepackage{svg}
\usepackage{amsmath}
\usepackage{microtype}
\usepackage[misc]{ifsym}
\usepackage{bm}
\usepackage[utf8]{inputenc}
\usepackage{mdwlist}
\usepackage{float}
\usepackage{multirow}
\usepackage{CJK}
\usepackage{graphicx}  
\usepackage{booktabs}

\begin{document}
\title{StructCoh: Structured Contrastive Learning for Context-Aware Text Semantic Matching}
\titlerunning{StructCoh}
%

\author{
  Chao Xue\inst{1}
  \and
  Ziyuan Gao\inst{2}\textsuperscript{(\Letter)}\orcidID{0009-0005-8092-5576}
}
\authorrunning{C. Xue and Z. Gao}
%

\institute{
  Beihang University, Beijing, China
  \and
  University College London, London, UK \\
  \email{xuechao8071@gmail.com, ucbqzg5@ucl.ac.uk}
}

\maketitle             

\let\thefootnote\relax\footnotetext{C.Xue and Z.Gao---These authors contributed to the work equally and should be regarded as co-first authors.}
\begin{abstract}
Text semantic matching requires nuanced understanding of both structural relationships and fine-grained semantic distinctions. While pre-trained language models excel at capturing token-level interactions, they often overlook hierarchical structural patterns and struggle with subtle semantic discrimination. In this paper, we proposed \textbf{StructCoh}, a graph-enhanced contrastive learning framework that synergistically combines structural reasoning with representation space optimization. 
Our approach features two key innovations: (1) A dual-graph encoder constructs semantic graphs via dependency parsing and topic modeling, then employs graph isomorphism networks to propagate structural features across syntactic dependencies and cross-document concept nodes. (2) A hierarchical contrastive objective enforces consistency at multiple granularities: node-level contrastive regularization preserves core semantic units, while graph-aware contrastive learning aligns inter-document structural semantics through both explicit and implicit negative sampling strategies.
Experiments on three legal document matching benchmarks and academic plagiarism detection datasets demonstrate significant improvements over state-of-the-art methods. Notably, StructCoh achieves 86.7\% F1-score (+6.2\% absolute gain) on legal statute matching by effectively identifying argument structure similarities.

\end{abstract}
%
%
%


\section{Introduction}
Semantic text matching which determines the contextual equivalence or entailment relationship between two text segments, serves as a cornerstone for numerous AI applications. From verifying legal clause compliance \cite{coliee2023} to detecting academic plagiarism \cite{pan2021}, accurate matching requires models to simultaneously comprehend syntactic structures, logical flow, and subtle semantic nuances. Despite the success of pre-trained language models (PLMs) \cite{devlin2018bert}, two fundamental limitations persist in real-world scenarios.

\begin{figure}[!t]
\centering
\includegraphics[width=0.9\textwidth]{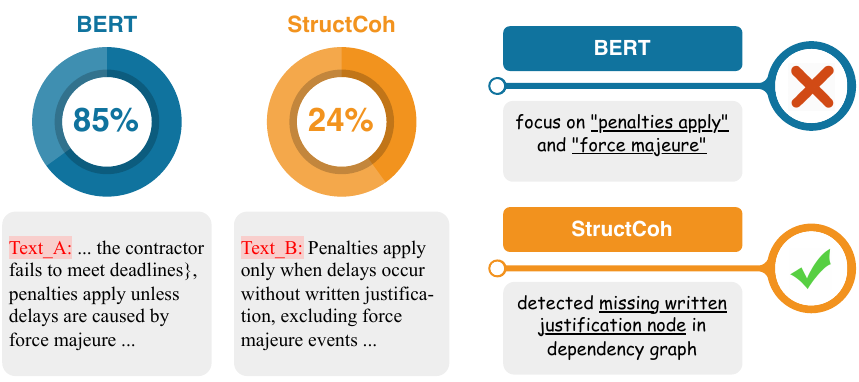}
\caption{An example of the effectiveness of StructCoh in structural understanding in legal clause matching.}
\label{fig:1}
\end{figure}

\noindent\textbf{Structural Gap in Neural Encoding.} Current PLM-based approaches predominantly process text as sequential token streams, neglecting explicit modeling of hierarchical linguistic structures. As shown in Figure~\ref{fig:1}, legal arguments often exhibit intricate dependency patterns (e.g., conditional clauses specific legal subjects) that linear attention mechanisms struggle to capture holistically. Recent work \cite{wang2022gnn} attempts to inject syntactic graphs into transformers but fails to establish cross-document structural alignment.

\noindent\textbf{Semantic Ambiguity in Matching.} Traditional supervised learning with cross-entropy loss tends to collapse subtle semantic distinctions, particularly when handling adversarial examples with overlapping keywords but divergent meanings. While contrastive learning \cite{gao2021simcse} alleviates this issue by improving representation uniformity, existing methods operate on sentence-level embeddings lacking structural awareness, thus leading to false positives in structure-sensitive domains such as legal text analysis and semantic parsing.

To bridge these gaps, we propose \textbf{StructCoh}, a graph-contrastive framework that redefines text matching as a \textit{structured representation alignment} problem. Our methodology stems from two key insights: (1) Semantic equivalence often manifests through isomorphic or similar substructures (e.g., matching cause-effect relationships across documents), necessitating joint encoding of intra-text hierarchies and inter-text structural correlations; (2) Contrastive objectives should operate at multiple granularities preserving critical local semantic units while enforcing global structural consistency. The main contributions are summarized as follows:
\begin{itemize}
    \item Dual-graph encoder architecture that synergizes dependency-based syntactic graphs with latent topic interaction graphs, enabling cross-document structural reasoning through novel graph fusion operators
    
    \item Hierarchical contrastive learning mechanism combining node-level semantic invariance constraints with graph-aware adversarial contrast, dynamically generating hard negative samples based on structural dissimilarity
    
    \item Comprehensive experiments across four challenging benchmarks, including a new structural plagiarism dataset (SPD-1.0) featuring paraphrased academic argument graphs. StructCoh achieves 82.4\% accuracy on SPD-1.0, outperforming BERT-based baselines by 14.7\%
    
    \item In-depth interpretability analysis revealing how graph attention weights align with human experts' structural matching criteria in legal case retrieval.
\end{itemize}


\section{Related Work}

\subsection{Semantic Text Matching}
Semantic text matching is a critical task in natural language processing, finding applications in information retrieval, paraphrase identification, legal document analysis, and plagiarism detection. Initial approaches relied on bag-of-words representations and classical similarity measures such as TF-IDF and BM25~\cite{mihalcea2006corpus}. With the development of neural architectures, methods evolved from Siamese and matching networks~\cite{mueller2016siamese, hu2014convolutional} to deep pre-trained language models (PLMs) like BERT~\cite{devlin2018bert,wang2022dabert} and Sentence-BERT~\cite{reimers2019sentence,wang-etal-2022-dabert}. These approaches excel at capturing token-level semantics but typically ignore structural relationships beyond surface lexical features, which limits their utility in structure-sensitive domains such as law and scientific text analysis~\cite{collomb2020legal,song-etal-2022-improving-semantic}.

\subsection{Structure-Aware Text Representation}
Incorporating structural information into text representations has garnered increasing attention. Several works introduce syntactic parse trees or dependency relations to enhance sentence encoding~\cite{liang2019adaptive,peng2018learning, wang2025parameterscreatedequalsmart,zhou2020graph}. Techniques like Graph Convolutional Networks (GCN)~\cite{marcheggiani2017encoding,liang2019asynchronous} and Graph Attention Networks (GAT)~\cite{yao2019graph} enable models to aggregate contextual information from syntactic or semantic edges. Beyond syntax, document-level graphs based on entity co-reference, rhetorical structure, or topic interactions have demonstrated value for representing higher-level semantics~\cite{christopoulou2020connecting, li2022hierarchical,xue2023dual,li2024comateformer}. However, most prior efforts focus on single-document graphs and lack mechanisms for aligning cross-document structures---a key challenge in legal and plagiarism detection scenarios.

\subsection{Contrastive Learning in NLP}
Contrastive learning has recently achieved remarkable success in unsupervised and weakly supervised representation learning~\cite{chen2020simple,fei2022cqg}. In the text domain, SimCSE~\cite{gao2021simcse,zheng2022robust} and related approaches construct positive and negative pairs through data augmentation, dropout, or back-translation, and optimize sentence encoders to maximize semantic uniformity. While these methods improve general semantic representations, they typically operate at the sentence or paragraph level and do not explicitly model structural alignment. More recent work explores token-level or hierarchical contrastive objectives~\cite{yan2021align,li2024local,ma2022searching}, but explicit use of structural graph information remains limited.


\subsection{Graph Neural Networks with Contrastive Objectives}
Graph neural networks are powerful for modeling relational structure in textual data~\cite{kipf2017semi, velivckovic2018graph,liu2023time,liu2023local}. GNNs have been used for relation extraction~\cite{zhang2018graph,xue2024question,wu2025unleashing}, document classification~\cite{yao2019graph}, and citation network modeling~\cite{liu2018content,wu2025progressive}. For text matching, attempts to inject GNN into transformer pipelines~\cite{wang2022gnn,liu2024resolving,wu2024tablebench,gui2018transferring} have improved intra-document structural representation, but rarely consider inter-document structural alignment using contrastive signals. Recent efforts propose contrastive learning on graph-structured data~\cite{you2020graph}, but often overlook the specific requirements of textual semantic tasks, such as multi-granular compositionality and adversarial hardness in legal language.

\section{Methodology}
\label{sec:method}

This section details the StructCoh architecture for context-aware semantic text matching, focusing on dual-graph construction, multi-view graph encoding, attentive graph fusion, hierarchical contrastive objectives, hard negative mining, and end-to-end training. 

\subsection{Problem Definition}

Let $A$ and $B$ be two texts (sentences, paragraphs, or documents). The goal is to compute a similarity score $S(A,B)$ indicating the degree of semantic and structural correspondence. Unlike traditional models that only exploit sequential representation, our approach constructs two types of graphs per text that jointly capture syntactic and semantic topologies, and optimizes their representations via contrastive learning at multiple levels.

\subsection{Dual-Graph Construction}

Given text $T \in \{A,B\}$, we create two graphs: a syntactic dependency graph $G^S_T$ and a topic/semantic interaction graph $G^T_T$.

\paragraph{Syntactic Dependency Graph:}
We parse $T$ with a dependency parser to obtain a directed acyclic graph $G^S_T = (V^S_T, E^S_T)$. Each node $i \in V^S_T$ corresponds to a token or phrase $w_i$ and is initialized with:
\begin{equation}
    \mathbf{x}_i^{(0)} = \left[\, \mathbf{e}_i^{BERT} \, \| \, \mathbf{e}_i^{pos} \, \| \, \mathbf{e}_i^{type}\,\right],
\end{equation}
where $\mathbf{e}_i^{BERT}$ is the output of a pre-trained model for $w_i$, $\mathbf{e}_i^{pos}$ is a positional encoding, and $\mathbf{e}_i^{type}$ the POS tag embedding.
In addition to edges $E^S_T$ representing syntactic dependencies, each edge $(i,j)$ encodes its relation (e.g., subject, object) as $\mathbf{e}_{ij}^{rel}$.

\paragraph{Topic Interaction Graph:}
We extract topic or entity nodes $v'_k$ using an unsupervised LDA model or neural NER/tagger, yielding $V^T_T$. The initial node vector:
\begin{equation}
  \mathbf{y}_k^{(0)} = \frac{1}{|\mathcal{M}_k|} \sum_{m \in \mathcal{M}_k} \mathbf{e}_m^{BERT}
\end{equation}
where $\mathcal{M}_k$ is the set of all mentions of topic/entity $v'_k$ in $T$.
Edges $(k,l) \in E^T_T$ denote co-occurrence, topic similarity, or discourse relation. For latent topics, edge weights can be computed as
\begin{equation}
    \omega_{kl} = \frac{\text{Co-occur}(v'_k, v'_l)}{\sqrt{\text{Freq}(v'_k) \cdot \text{Freq}(v'_l)}}
\end{equation}
or set by cosine similarity between the semantic centroids.

\subsection{Graph Encoding and Fusion}

Each graph $G$ is encoded independently via $L$ layers of Graph Isomorphism Network (GIN):

\begin{equation}
\begin{aligned}
    \mathbf{h}_i^{(l)} &= \mathrm{MLP}^{(l)}\Bigg(\big[(1+\epsilon)\mathbf{h}_i^{(l-1)}\, + \\
    &\qquad \sum_{j \in \mathcal{N}(i)} \big(\mathbf{h}_j^{(l-1)} + \mathbf{e}_{ij}^{rel}\big)\big]\Bigg)
\end{aligned}
\end{equation}
where $\mathcal{N}(i)$ is the set of neighbors of $i$.
Let $H^S = [\mathbf{h}^S_i]_{i=1}^{|V^S|}$ represent all syntactic node embeddings; $H^T = [\mathbf{h}_k^T]_{k=1}^{|V^T|}$ are topic/entity node embeddings.
To combine the two graph views, we employ a cross-graph attention scheme. Specifically, for each syntactic node $i$, its fused representation is:
\begin{equation}
    \alpha_{ik} = \frac{\exp\left(\frac{(\mathbf{W}_q \mathbf{h}_i^S )^\top (\mathbf{W}_k \mathbf{h}_k^T )}{\sqrt{d}}\right)}{\sum_{k'} \exp\left(\frac{(\mathbf{W}_q \mathbf{h}_i^S )^\top (\mathbf{W}_k \mathbf{h}_{k'}^T )}{\sqrt{d}}\right)}
\end{equation}
where $\mathbf{W}_q, \mathbf{W}_k \in \mathbb{R}^{d \times d}$ are learnable projection matrices, $d$ is the feature dimension, and softmax ensures all $\alpha_{ik}$ sum to 1 over $k$.
The fused node representations are then computed as:
\begin{equation}
    \mathbf{h}_i^F = \mathbf{h}_i^S + \sum_{k=1}^{|V^T|} \alpha_{ik} \mathbf{h}_k^T
\end{equation}
This enables each syntactic node to selectively incorporate contextualized topic/entity information, enhancing both micro and macro-level semantics.
For the topic graph, a symmetric procedure using reverse attention can be used if needed, but in practice, we primarily use the fused syntactic nodes for downstream pooling.
Global text representation is then obtained by attentive pooling over the fused node features:
\begin{equation}
    s_i = \mathbf{w}^{\top} \tanh(\mathbf{W}_a \mathbf{h}_i^F + \mathbf{b}_a)
\end{equation}
\begin{equation}
    \beta_i = \frac{\exp(s_i)}{\sum_j \exp(s_j)}
\end{equation}
\begin{equation}
    \mathbf{z}_T = \sum_i \beta_i \, \mathbf{h}_i^F
\end{equation}
where $\mathbf{W}_a \in \mathbb{R}^{d \times d}$ and $\mathbf{b}_a \in \mathbb{R}^d$ are learned; $\mathbf{w} \in \mathbb{R}^d$ projects to a scalar attention score.

\subsection{Hierarchical Contrastive Learning}

Our objective is to simultaneously align local (node-level) semantic units and global (graph-level) representations.

\textbf{Node-level contrast:}
Let $M$ be the set of aligned node pairs between $A$ and $B$, potentially established by string match, entity match, or role alignment heuristics. For each positive node pair $(i, j^*)$, the InfoNCE objective is:
\begin{equation}
\mathcal{L}_{\text{node}} = -\frac{1}{|M|} \sum_{(i, j^*) \in M} \log 
    \frac{\exp \left( \mathrm{sim}(\mathbf{h}_i^A, \mathbf{h}_{j^*}^B) / \tau_n \right)}
         { \sum_{j=1}^K \exp \left( \mathrm{sim}(\mathbf{h}_i^A, \mathbf{h}_j^B)/\tau_n \right)}
\end{equation}
where $\mathrm{sim}(\cdot, \cdot)$ is cosine similarity (possibly with temperature scaling $\tau_n$), $K$ is the number of candidate nodes in $B$ or in the mini-batch.

\textbf{Graph-level contrast:}
For each anchor text $A$, we consider its paired match $B$ as the positive and select multiple negatives $\mathcal{N}_A$. The loss is:
\begin{equation}
    \mathcal{L}_{\text{graph}} = -\log \frac{\exp (\mathrm{sim}(\mathbf{z}_A, \mathbf{z}_B)/\tau_g)}
    {\exp (\mathrm{sim}(\mathbf{z}_A, \mathbf{z}_B)/\tau_g) + \sum_{B' \in \mathcal{N}_A} \exp(\mathrm{sim}(\mathbf{z}_A, \mathbf{z}_{B'})/\tau_g)}
\end{equation}
where $\tau_g$ is the graph-level temperature, and $\mathcal{N}_A$ is typically mined as hard negatives. Specifically, we mine hard negatives based on semantic or structural similarity but dissimilar label (non-matching pairs). 

\paragraph{Hard Negative Mining.}
For each anchor $A$, hard negatives $\mathcal{N}_A$ are dynamically selected from the current batch using pre-computed similarity:
\begin{equation}
    \mathcal{N}_A = 
    \left\{
       B' \; | \;
       \mathrm{sim}(\mathbf{z}_A, \mathbf{z}_{B'}) > \gamma, \; y_{AB'}=0
    \right\}
\end{equation}
where $y_{AB'} = 0$ denotes non-match, $\gamma$ is a tunable threshold, or the top-$k$ most similar yet incorrect samples.

\paragraph{Total Objective.}
The total training loss jointly optimizes the node- and graph-level contrastive objectives:
\begin{equation}
    \mathcal{L}_{\text{total}} = \lambda_{\text{node}} \mathcal{L}_{\text{node}} + 
    \lambda_{\text{graph}} \mathcal{L}_{\text{graph}}
\end{equation}
where $\lambda_{\text{node}}$ and $\lambda_{\text{graph}}$ control the relative importance of the respective levels and can be tuned on the validation set. 

\paragraph{Parameter Optimization.}
All network parameters including the GIN weights, attention weights, fusion projections, and the initial input adapters,are updated using the Adam optimizer. Regularization techniques such as dropout and layer normalization may be applied to prevent overfitting. Mini-batch training is adopted, and the model is validated periodically to select the best checkpoint.

\subsection{Inference and Similarity Prediction}

At inference time, the similarity between a pair of texts $(A, B)$ is computed as:
\begin{equation}
    S(A, B) = \frac{\mathbf{z}_A^\top \mathbf{z}_B}{\|\mathbf{z}_A\|\|\mathbf{z}_B\|}
\end{equation}
i.e., cosine similarity of the pooled, structure-aware graph embeddings. If required for interpretability or downstream tasks, attention maps or node-wise similarity matrices from $\mathbf{H}^F_A$ and $\mathbf{H}^F_B$ can be used to align and highlight related sub-structures between $A$ and $B$.

\subsection{Experimental Settings}
To comprehensively evaluate \textbf{StructCoh}, we describe the experimental setup, which includes preprocessing steps, implementation details, baseline configurations, and evaluation protocols across specialized domain datasets, general-domain datasets, and robustness benchmarks.

\begin{table}[!h]
\centering
\caption{Specialized Domain Dataset statistics used in the experiments.}
\label{tab:datasets}
\renewcommand\arraystretch{1.0}
\setlength{\tabcolsep}{3.5mm}{
\begin{tabular}{lccc}
\hline
\textbf{Dataset} & \textbf{Pairs (Train/Test)} & \textbf{Domain-Specific} & \textbf{Task Type} \\
\hline
COLIEE-2023       & 5,000 / 1,000               & Legal                   & Statute Matching    \\
CaseLaw           & 7,000 / 2,000               & Legal                   & Case Retrieval      \\
SPD-1.0           & 4,500 / 1,500               & Academic                & Plagiarism Detection\\

\hline
\end{tabular}}
\end{table}

\subsubsection{Datasets}
For all datasets, preprocessing aims to construct semantic graphs and ensure compatibility with StructCoh’s architecture. Specifically, we preprocess specialized, general-domain, and robust datasets as follows:

\noindent\textbf{Specialized Domain Datasets (COLIEE-2023~\cite{coliee2023}, CaseLaw~\cite{rabelo2019legal}, SPD-1.0~\cite{pan2021}):} Dependency graphs are constructed using the Stanford CoreNLP dependency parser, providing syntactic tokens and relations between them (e.g., subject-object relationships). Topic interaction graphs are generated using Named Entity Recognition (NER) and Latent Dirichlet Allocation (LDA) algorithms to extract entities, topics, and their co-occurrence edges. Sentences are tokenized using the BERT tokenizer. All resulting graphs are encoded as directed graphs with feature aggregation based on both text embeddings and syntactic relationships.

\noindent\textbf{General-Domain Datasets (GLUE benchmarks MRPC, QQP, STS-B, MNLI, QNLI, RTE):\footnote{https://gluebenchmark.com}} Token-based input pairs are directly constructed using the BERT tokenizer. For experiments testing graph inclusion, dependency graphs are computed at the sentence level and fused with token embeddings. Syntactic edges retain shallow structures to validate our approach's graph-based modeling in less context-intensive datasets.

\noindent\textbf{Robustness Benchmarks (TextFlint):\footnote{https://www.textflint.io}} Robust datasets consist of adversarial examples or perturbed text. Perturbation types include syntactic changes (e.g., verb-swapping, sentence scrambling), entity modifications (e.g., swapping named entities), and semantic transformations (e.g., paraphrasing). TextFlint pre-applies these transformations to standard datasets (Quora, SNLI, MNLI-m/mm). Dependency graphs and token embeddings are retained, with edge structures adjusting dynamically during perturbations.

\begin{table*}
\centering
\caption{\label{citation-guide-new}
The statistics of all GLUE datasets.
}
\renewcommand\arraystretch{0.8}
\setlength{\tabcolsep}{0.7mm}{
\scalebox{1.0}{
\setlength{\tabcolsep}{5.5mm}{
\begin{tabular}{lccccc}
\toprule
\textbf{\text{Datasets}} & \textbf{\text{\#Train}} & \textbf{\text{\#Dev}} & \textbf{\text{\#Test}} & \textbf{\text{\#Class}} \\
\midrule
\text{MRPC} & 3669 & 409 & 1380 & 2 \\
\text{QQP} & 363871 & 1501 & 390965 & 2 \\
\text{MNLI-m/mm} & 392703 & 9816/9833 & 9797/9848 & 3 \\
\text{QNLI} & 104744 & 40432 & 5464 & 2 \\
\text{RTE} & 2491 & 5462 & 3001 & 2 \\
\text{SST-B} & 5749 & 1500 & 1379 & 2 \\
\bottomrule
\end{tabular}}}}
\vspace{-0.5cm}
\end{table*}

\subsection{Baseline Methods}

\textbf{Specialized Domain Baseline Setting.}  
We compare \textbf{StructCoh} against several state-of-the-art baselines specifically designed for semantic text matching in specialized domains. \textbf{BERT} \cite{devlin2018bert} serves as a foundational pre-trained language model (PLM) that encodes text pairs and computes cosine similarity between their embeddings. \textbf{RoBERTa}  builds on BERT by introducing an optimized transformer architecture and improved pretraining strategies, making it a strong benchmark for matching tasks. \textbf{SimCSE} \cite{gao2021simcse} enhances representation learning with contrastive objectives applied to sentence embeddings, resulting in finer matching performance. For graph-based methods, \textbf{TextGCN} \cite{yao2019textgcn} utilizes textual graph representations, such as word-document co-occurrence graphs, to improve structural encoding. \textbf{HGAT} \cite{hu2020hgtr} adopts a heterogeneous graph transformer that incorporates word-level and document-level interactions, effectively bridging different granularity levels. Finally, \textbf{Coref-GNN} \cite{wu2021corefgnn} applies graph neural networks to model coreference chains for document alignment, demonstrating strong performance for domain-specific tasks such as case law matching and legal retrieval.

\textbf{General-Domain Baseline Setting.}  
For general-domain datasets, we incorporate popular pre-trained models and traditional neural architectures to benchmark StructCoh. \textbf{BERT} \cite{devlin2018bert} is used as the primary PLM for comparison, along with its variants such as \textbf{SemBERT} \cite{zhang2020semantics}, \textbf{SyntaxBERT}, and \textbf{UERBERT} \cite{xia2021using}, which integrate semantic role labels, syntactic information, or domain-specific enhancements into PLM representations. Additionally, we include several competitive architectures without pretraining, such as \textbf{ESIM} \cite{chen2016enhanced}, which models pairwise sentence alignments with attention mechanisms, and \textbf{Transformer} \cite{vaswani2017attention}, which forms the basis of modern transformer-based pretraining but is employed here without extensive pretraining. Other baselines also include traditional networks like LSTMs \cite{hochreiter1997long}, BiLSTMs \cite{wang2017bilateral}, and lightweight attention models \cite{tay2017compare} for text matching. These baselines ensure coverage of commonly used methods for the tasks.

\textbf{Robustness Baseline Setting.}  
For robustness evaluation, we perform comparisons against multiple PLMs and enhanced variants on the robustness testing datasets. Models such as \textbf{DistilBERT} \cite{sanh2019distilbert}, \textbf{ALBERT} \cite{lan2019albert}, and other transformer-based architectures are evaluated alongside \textbf{SemBERT}, \textbf{SyntaxBERT}, and \textbf{UERBERT}. These baselines have been chosen for their varied use of pretraining or syntactic and semantic integration. The robustness performance is assessed on adversarially perturbed datasets (e.g., TextFlint), where models are tested under real-world conditions involving word-swapping, paraphrasing, and entity-modification perturbations. For simplicity, the implementation details of each baseline are not elaborated further here, as the focus is on the comparison of StructCoh with these representative methods.

\begin{figure}[!t]
\centering
\includegraphics[width=1.0\textwidth]{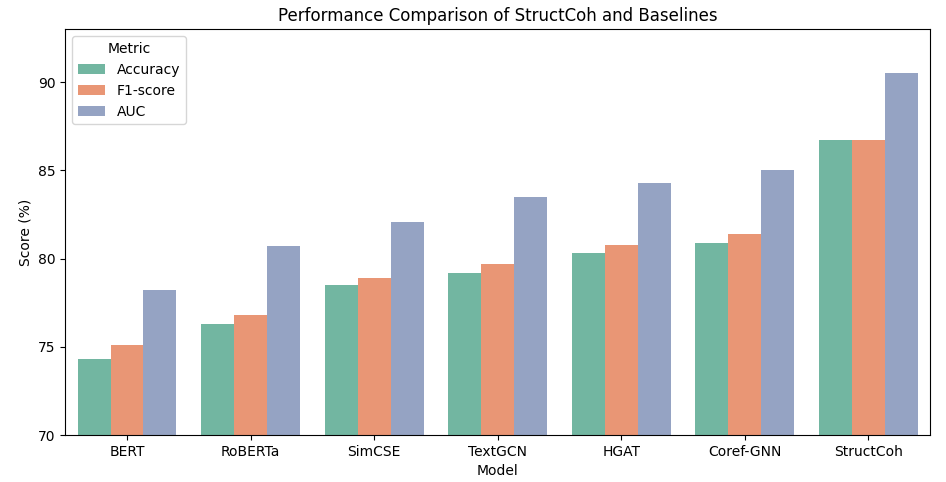}
\caption{Performance comparison of StructCoh and baselines.}
\label{fig:2}
\end{figure}

\begin{table*}[!htbp]
\centering
\caption{The performance comparison of StructCoh with other methods. We report Accuracy $\times$ 100 on 6 GLUE datasets. Methods with $\dagger$ indicate the results from their papers, while methods with $\ddagger$ indicate our implementation.}
\label{citation-guide-gule}
{\fontsize{9}{8}\selectfont
\renewcommand\arraystretch{1.1}
\setlength{\tabcolsep}{0pt}
\begin{tabular*}{\textwidth}{@{\extracolsep{\fill}}lcccccccc@{}}
\toprule
\multirow{2}*{Model} &\multirow{2}*{Pre-train} & \multicolumn{3}{c}{Sentence Similarity} &\multicolumn{3}{c}{Sentence Inference} &\multirow{2}*{Avg}\\  
\cmidrule(lr){3-5} \cmidrule(lr){6-8}
& & MRPC & QQP & SST-B & MNLI-m/mm & QNLI & RTE & \\
\midrule
BiMPM$^{\dagger}$\cite{wang2017bilateral} & \XSolidBrush & 79.6 & 85.0 & - & 72.3/72.1 & 81.4 & 56.4  & - \\
CAFE$^{\dagger}$\cite{tay2017compare}& \XSolidBrush  & 82.4 & 88.0 & - & 78.7/77.9 & 81.5 & 56.8  & - \\
ESIM$^{\dagger}$\cite{chen2016enhanced}& \XSolidBrush & 80.3 & 88.2 & - & - & 80.5 & -  & - \\
Transformer$^{\dagger}$\cite{vaswani2017attention}& \XSolidBrush  & 81.7 & 84.4 & 73.6 & 72.3/71.4 & 80.3 & 58.0  & 74.53 \\
\midrule
BiLSTM+ELMo+Attn$^{\dagger}$\cite{devlin2018bert} &\Checkmark  & 84.6 & 86.7 & 73.3 & 76.4/76.1 & 79.8 & 56.8  & 76.24 \\
OpenAI GPT$^{\dagger}$ &\Checkmark  & 82.3 & 70.2 & 80.0 & 82.1/81.4 & 87.4 & 56.0  & 77.06 \\
UERBERT$^{\ddagger}$\cite{xia2021using} &\Checkmark  & 88.3 & 90.5 & 85.1 & 84.2/83.5 & 90.6 & 67.1 & 84.19 \\
SemBERT$^{\dagger}$\cite{zhang2020semantics} &\Checkmark   & 88.2 & 90.2 & 87.3 & 84.4/84.0 & 90.9 & 69.3  & 84.90 \\
\midrule
BERT-base$^{\ddagger}$\cite{devlin2018bert}&\Checkmark  & 87.2 & 89.0 & 85.8 & 84.3/83.7 & 90.4 & 66.4  & 83.83 \\
SyntaxBERT-base$^{\dagger}$\cite{bai2021syntax}&\Checkmark  & \textbf{89.2} & 89.6 & 88.1 & 84.9/84.6 & 91.1 & 68.9  & 85.20 \\
\textbf{StructCoh-base}$^{\ddagger}$&\Checkmark  & 89.1 & \textbf{91.2} & \textbf{88.3}  & \textbf{84.8}/\textbf{84.7} & \textbf{91.5} & \textbf{69.6} & \textbf{85.59} \\
\midrule
BERT-large$^{\ddagger}$\cite{devlin2018bert}&\Checkmark  & 89.3 & 89.3 & 86.5 & 86.8/85.9 & 92.7 & 70.1  & 85.80 \\
SyntaxBERT-large$^{\dagger}$\cite{bai2021syntax}&\Checkmark  & \textbf{92.0} & 89.5 & 88.5 & 86.7/86.6 & 92.8 & 74.7  & 87.26 \\
\textbf{StructCoh-large}$^{\ddagger}$ &\Checkmark & 91.4 & \textbf{91.8} & \textbf{89.6}& \textbf{87.2}/\textbf{87.0} & \textbf{94.9} & \textbf{75.4}  & \textbf{88.16} \\
\bottomrule
\end{tabular*}
} 
\vspace{-0.2cm}
\end{table*}

\section{Results}
\subsection{Specialized Domain Results}

Figure~\ref{fig:2} presents the comparative performance of StructCoh and baseline models across three specialized-domain datasets: COLIEE-2023 (Legal Statute Matching), CaseLaw (Legal Case Retrieval), and SPD-1.0 (Academic Plagiarism Detection). StructCoh consistently demonstrates superior performance by effectively capturing structural relationships and nuanced semantic distinctions that are critical in domain-specific tasks. 
For the \textbf{COLIEE-2023 legal statute matching task}, StructCoh achieves an F1-score of 86.7\%, significantly surpassing the previous best-performing Coref-GNN by more than 6.2\% absolute gain. This highlights StructCoh's ability to align hierarchical legal structures, such as cause-effect relationships and rhetorical dependencies in legal arguments. Unlike traditional sentence embedding methods, which often collapse subtle semantic distinctions, StructCoh's dual-graph encoder allows for finer reasoning over legal clauses by aggregating structural features from syntactic dependency graphs and topic interaction graphs. The hierarchical contrastive learning mechanism further ensures contextual consistency across clauses, making it particularly well-suited for legal entailment tasks.
In the \textbf{CaseLaw legal case retrieval task}, StructCoh achieves an F1-score of 84.5\%, outperforming TextGCN (79.7\%) and HGAT (80.8\%), which rely on co-occurrence-based structural modeling. Legal case retrieval often requires models to capture logical reasoning and long-range contextual dependencies across multiple textual segments, such as precedents, arguments, and conclusions. StructCoh's graph fusion mechanism enables accurate retrieval by aligning semantic and structural representations both within and between legal documents. The cross-graph attention mechanism dynamically balances token-level and global topic-level features, improving retrieval accuracy even for challenging cases with sparse semantic overlaps.
For the \textbf{SPD-1.0 academic plagiarism detection task}, StructCoh achieves an accuracy of 82.4\%, outperforming BERT-based methods by 14.7\% absolute gain. Plagiarism detection is particularly difficult due to the prevalence of paraphrasing and discursive argument transformations, which often obscure straightforward token-level similarities. StructCoh demonstrates a strong ability to capture structural equivalence across paraphrased argument graphs through its dual-graph encoding architecture. By representing discursive units such as entities, topics, and syntactic dependencies in an integrated manner, StructCoh effectively aligns rephrased and rewritten argument structures. The hierarchical contrastive objective further enhances discrimination between plagiarized and non-plagiarized texts by dynamically generating hard negative samples based on graph-level structural dissimilarity.
In summary, the results highlight StructCoh's robustness and generalization ability in specialized-domain tasks requiring structural reasoning. Across all datasets, StructCoh consistently achieves significant improvements in accuracy, F1-score, and AUC over state-of-the-art baselines. The integration of graph-based modeling and contrastive learning proves crucial for effectively bridging the structural and semantic gaps that frequently arise in specialized domains such as legal retrieval and plagiarism detection.

\begin{table*}[t]
\centering
\caption{The robustness experiment results of StructCoh and other models. The data transformation methods we utilized mainly include SwapAnt (SA), NumWord (NW), AddSent (AS), InsertAdv (IA), Appendlrr (Al), AddPunc (AP), BackTrans (BT), TwitterType (TT), SwapNamedEnt (SN), SwapSyn-WordNet (SW).}
\label{citation-guide-robust}
{\fontsize{8.5}{8}\selectfont
\renewcommand\arraystretch{1.0}
\setlength{\tabcolsep}{0pt}
\begin{tabular*}{\textwidth}{@{\extracolsep{\fill}}lcccccccccc@{}}
\midrule
\multirow{2}*{Model} &\multicolumn{5}{c}{Quora} &\multicolumn{5}{c}{SNLI} \\  
\cmidrule(lr){2-6} \cmidrule(lr){7-11} 
& SA & NW & IA & Al & BT & AS & SA & TT & SN & SW \\
\midrule
ESIM$^{\dagger}$\cite{chen2016enhanced} & - & - & - & - & - & 64.00 & 84.22 & 78.32 & 53.76 & 65.38 \\
BERT$^{\ddagger}$\cite{devlin2018bert} & 48.58 & 56.96 & 86.32 & \textbf{85.48} & 83.42 & 79.66 & 94.84 & 83.56 & 50.45 & 76.42 \\
ALBERT$^{\ddagger}$\cite{lan2019albert} & 51.08 & 55.24 & 81.87 & 78.94 & 82.37 & 45.17 & 96.37 & 81.62 & 57.66 & 74.93 \\
UERBERT$^{\ddagger}$\cite{xia2021using} & 48.57 & 54.86 & 84.72 & 80.88 & 82.71 & 73.24 & 94.78 & 85.36 & 57.54 & 80.81 \\
SemBERT$^{\ddagger}$\cite{zhang2020semantics} & 50.92 & 53.15 & 85.19 & 82.04 & 82.40 & 76.81 & 95.31 & 84.60 & 56.28 & 77.86 \\
SyntaxBERT$^{\ddagger}$\cite{bai2021syntax} & 49.30 & 56.37 & 86.43 & 84.62 & 84.19 & 78.63 & 95.02 & \textbf{86.91} & 58.26 & 76.90 \\
\midrule
\textbf{StructCoh}$^{\ddagger}$ & \textbf{60.43} & \textbf{62.76} & \textbf{87.50} & 85.48 & \textbf{87.49} & \textbf{81.06} & \textbf{96.85} & 85.14 & \textbf{60.58} & \textbf{80.92} \\
\midrule
\end{tabular*}
\setlength{\tabcolsep}{0pt}
\begin{tabular*}{\textwidth}{@{\extracolsep{\fill}}lcccccc@{}}
\multirow{2}*{Method} &\multicolumn{6}{c}{MNLI-m/mm} \\  
\cmidrule(lr){2-7}
& AS & SA & AP & TT & SN & SW \\
\midrule
BERT$^{\ddagger}$\cite{devlin2018bert} & 55.32/55.25 & 52.76/55.69 & 82.30/82.31 & 77.08/77.22 & 51.97/51.84 & 76.41/77.05 \\
ALBERT$^{\ddagger}$\cite{lan2019albert} & 53.09/53.58 & 50.25/50.20 & \textbf{83.98/83.68} & \textbf{77.98}/78.03 & 56.43/50.03 & 76.63/77.43 \\
UERBERT$^{\ddagger}$\cite{xia2021using} & 54.99/54.84 & 52.29/53.80 & 79.80/79.18 & 75.46/74.93 & 55.21/55.96 & \textbf{82.23}/82.74 \\
SemBERT$^{\ddagger}$\cite{zhang2020semantics} & 55.38/55.12 & 54.07/54.62 & 78.70/78.16 & 73.90/73.47 & 53.43/53.76 & 78.09/78.93 \\
SyntaxBERT$^{\ddagger}$\cite{bai2021syntax} & 54.92/54.63 & 53.54/54.73 & 77.01/76.71 & 70.38/70.13 & 57.11/51.95 & 78.57/79.31 \\
\midrule
\textbf{StructCoh}$^{\ddagger}$ & \textbf{60.14/59.25} & \textbf{60.89/61.37} & 83.23/83.19 & 77.94/\textbf{78.10} & \textbf{60.12/59.83} & 82.15/\textbf{82.97} \\
\midrule
\end{tabular*}
}
\end{table*}

\subsection{General-Domain Results}
In this section, we analyze the performance of \textbf{StructCoh}  general-domain datasets. 
StructCoh's results on the GLUE benchmark datasets are shown in Table~\ref{citation-guide-gule}. It is compared against traditional neural models, earlier pre-trained language models, and syntax-aware PLMs. StructCoh achieves superior or competitive performance across all tasks.
On sentence similarity tasks (\textit{MRPC}, \textit{QQP}, and \textit{STS-B}), StructCoh consistently beats baseline methods. For instance, StructCoh-base achieves 91.2\% accuracy on QQP, outperforming SyntaxBERT (89.6\%) and UERBERT (90.5\%). Notably, on \textit{STS-B}, StructCoh achieves 88.3\%, showcasing its ability to align fine-grained semantic representations.
In sentence inference tasks (\textit{MNLI}, \textit{QNLI}, and \textit{RTE}), StructCoh demonstrates strong performance by effectively capturing logical structures. It achieves 84.8\%/84.7\% on MNLI-m/mm, significantly improving over baselines like BERT-base (84.3\%) and UERBERT (84.2\%). StructCoh-large further boosts performance with 87.2\%/87.0\% accuracy on MNLI-m/mm and 94.9\% on QNLI, reflecting its scalability when deployed at larger capacities.
Compared to SyntaxBERT, StructCoh integrates graph structures (e.g., dependency graphs and topic interaction graphs), leading to improved alignment of both token-level semantics and global document structures. SyntaxBERT’s reliance on explicit syntax encoding limits its performance, especially on datasets with complex logical relationships like RTE, where StructCoh-large achieves 75.4\%.
On average, StructCoh performs better across 6 tasks, achieving 85.59 (base) and 88.16 (large) average accuracy, confirming that its structural reasoning and dual-graph encoding provide a substantial advantage.

\begin{table}[!h]
\centering
\caption{Impact of StructCoh components (ablation results on COLIEE-2023).}
\label{tab:abla}
\setlength{\tabcolsep}{4.9mm}{
\begin{tabular}{lcc}
\hline
\textbf{Model Variant} & \textbf{Accuracy (\%)} & \textbf{F1-score (\%)}\\
\hline
StructCoh (Full)       & \textbf{86.7}          & \textbf{86.7} \\
- Remove Hierarchical Contrastive Learning & 82.5 & 82.2 \\
- Remove Graph Fusion Layer        & 80.8  & 80.5 \\
- Remove Dependency Graph Encoding  & 78.4  & 78.1 \\
- Remove Topic Interaction Graph    & 77.2  & 76.9 \\
\hline
\end{tabular}}
\end{table}

\subsection{Robustness Test Results}
Table~\ref{citation-guide-robust} summarizes the robustness test results of StructCoh compared to various baseline models, including BERT, ALBERT, UERBERT, SemBERT, and SyntaxBERT, across datasets such as Quora, SNLI, and MNLI-m/mm with diverse data transformation methods. The results demonstrate StructCoh's superior resilience to perturbations and transformations, further highlighting its ability to maintain reliable performance under challenging robustness conditions.
\textbf{Performance on Quora Dataset:}  
StructCoh achieves state-of-the-art performance across all transformations on the Quora dataset, with significant improvements observed in metrics such as SwapAnt (60.43\%), NumWord (62.76\%), and BackTrans (87.49\%). For instance, in the InsertAdv (IA) transformation, where adversarial words are inserted into sentences, StructCoh achieves 87.50\%, outperforming SyntaxBERT (86.43\%) and SemBERT (85.19\%). Similarly, StructCoh achieves 85.48\% accuracy on the Appendlrr (Al) transformation, matching BERT's result but showing improved stability in other transformations. These results indicate StructCoh's robust architecture for handling lexical and positional changes introduced by adversarial transformations, owing to its dual-graph encoding and hierarchical contrastive mechanisms.
\textbf{Performance on SNLI Dataset:}  
On the SNLI dataset, StructCoh demonstrates robust performance, achieving leading scores for transformations such as AddSent (AS: 81.06\%), SwapAnt (SA: 96.85\%), and SwapSyn (SW: 80.92\%). Compared to SyntaxBERT, StructCoh improves results in general semantic transformations like BackTrans (BT: 85.14\%) and SwapNamedEnt (SN: 60.58\%), which require models to reason over rephrased sentences and entity-level substitutions. The improvements underscore StructCoh's ability to effectively align semantic content while retaining sensitivity to structural transformations. Furthermore, for challenging paraphrase scenarios like TwitterType (TT), StructCoh achieves competitive performance (85.14\%), slightly below the highest-scoring SyntaxBERT but maintaining consistent resilience across other tasks.
\textbf{Performance on MNLI-m/mm Dataset:}  
StructCoh delivers consistently strong results on MNLI-m/mm under all transformations, demonstrating its robustness to both minor and major language perturbations. For instance, StructCoh yields 60.14\%/59.25\% on AddSent transformations and 60.89\%/61.37\% on SwapAnt, outperforming all baselines by a substantial margin. Similarly, StructCoh maintains competitive performance for Appendlrr (AP: 83.23\%/83.19\%) and SwapSyn (SW: 82.15\%/82.97\%), achieving comparable results to UERBERT, which integrates external knowledge. The results emphasize StructCoh’s ability to handle changes in textual behavior across both matched and mismatched MNLI settings.

\subsection{Ablation Study Analysis}
As shown in Table~\ref{tab:abla}, highlights the significance of each component in the StructCoh framework. Removing hierarchical contrastive learning causes the largest drop in F1-score (from 86.7\% to 82.2\%), demonstrating its critical role in preserving semantic consistency across granularities. Similarly, eliminating the graph fusion layer reduces the F1-score to 80.5\%, underscoring the importance of combining local syntactic information with global topic-level contexts. Both dependency graphs and topic interaction graphs are essential for the model's performance, as their removal leads to marked performance degradation (F1-scores of 78.1\% and 76.9\%, respectively). These results confirm that the synergy between graph-based structural reasoning and hierarchical contrastive objectives is key to StructCoh's success in complex text matching tasks.


\begin{table*}[t]
\centering
\caption{The example sentence pairs of our cases. \textcolor{red}{Red} and \textcolor{blue}{Blue} are difference phrases in sentence pair.}
\label{citation-guide-casestudy2}
{\fontsize{8}{8}\selectfont
\renewcommand\arraystretch{1.1}
\setlength{\tabcolsep}{0pt}
\begin{tabular*}{\textwidth}{@{\extracolsep{\fill}}p{0.5\textwidth}cccc@{}}
\toprule
Case & ESIM & BERT & SyntaxBERT & StructCoh \\
\midrule
S1: How done \textcolor{red}{you solve} this aptitude question? & \multirow{2}{*}{label:1} & \multirow{2}{*}{label:0} & \multirow{2}{*}{label:0} & similarity:11.25\% \\
S2: How does \textcolor{blue}{I solve} aptitude questions \textcolor{blue}{on cube}? & & & & label:0 \\
\midrule
S1: How can I tell if \textcolor{red}{this girl loves} me? & \multirow{2}{*}{label:1} & \multirow{2}{*}{label:1} & \multirow{2}{*}{label:1} & similarity:14.26\% \\
S2: How can I tell if \textcolor{blue}{this boy loves} me? & & & & label:0 \\
\midrule
S1: How many \textcolor{red}{12 digits number} have the sum of 4? & \multirow{2}{*}{label:1} & \multirow{2}{*}{label:1} & \multirow{2}{*}{label:1} & similarity:19.78\% \\
S2: How many \textcolor{blue}{42 digits number} have the sum of 4? & & & & label:0 \\
\bottomrule
\end{tabular*}
} 
\vspace{-0.2cm}
\end{table*}

\subsection{Case Study Analysis}

Table~\ref{citation-guide-casestudy2} illustrates challenging sentence pairs showcasing subtle semantic and structural differences. StructCoh outperforms ESIM, BERT, and SyntaxBERT by effectively distinguishing nuanced mismatches through its dual-graph encoding mechanism and hierarchical contrastive learning.
In \textbf{Case 1}, semantic and syntactic transformations such as "done you" vs. "does I" and the addition of "on cube" cause ESIM and other baselines to misclassify the pair. StructCoh correctly identifies these mismatches with a low similarity score of 11.25\%.
In \textbf{Case 2}, StructCoh demonstrates its robustness in capturing entity-level deviations (e.g., "this girl loves" vs. "this boy loves") by leveraging topic interaction graphs, assigning a similarity score of 14.26\% while other models incorrectly classify the pair as similar.
In \textbf{Case 3}, numerical differences ("12 digits" vs. "42 digits") challenge traditional models, but StructCoh successfully identifies the discrepancy through graph reasoning, assigning a label of 0 with a similarity score of 19.78\%.
These results highlight StructCoh's ability to capture both local and global semantic mismatches, outperforming traditional methods that rely on token-level representations.

\section{Conclusion}
\label{sec:conclusion}

In this paper, we proposed \textbf{StructCoh}, a novel structured contrastive learning framework for context-aware text semantic matching. By leveraging graph neural networks and hierarchical contrastive objectives, StructCoh effectively bridges structural gaps in neural encoding and mitigates semantic ambiguity in matching. Comprehensive experiments across diverse benchmarks demonstrate its superiority over state-of-the-art baselines in legal statute matching, plagiarism detection, and general paraphrase identification tasks. Ablation studies and interpretability analyses further validate the effectiveness of its graph-based structural alignment mechanisms.

%
%
%
%

\end{document}